# Towards an Autonomous Compost Turner: Current State of Research

Max Cichocki [1,*] , Eva Reitbauer [2] , Fabian Theurl [2], Christoph Schmied [2]

[1] Institute of Logistics Engineering (ITL), Graz University of Technology, Austria

[2] Institute of Geodesy, Graz University of Technology, Austria

* Correspondence: cichocki@tugraz.at

# 1  Introduction

Driven by climate change and the resulting need for action, global waste management has changed significantly in recent years. The European Union (EU) is therefore taking targeted action to promote composting and recycling [1]. This focus on a sustainable circular economy highlights the essential role of waste and discard within the overall life cycle of a product. Instead of being considered worthless assets, waste represents an integral contribution to the creation of new products. In the specific case of composting, organic waste is transformed into high-quality soils and substrates. However, the corresponding process to obtain quality products is often associated with challenging labor conditions. Therefore, within the present chapter an innovative project is presented, which investigates the development of a fully autonomous compost turner, with the aim of improving traditional composting methods and to overcome current challenges in the composting industry.

## 1.1  Composting as an integral part of the circular economy

Composting plays a crucial role in the circular economy, as it embodies the very fundamental idea of recovery, recycling and reuse of waste. [2] By transforming organic waste into valuable, high-quality products, composting contributes significantly to saving resources and reducing landfill waste.  Traditional linear economic models often take a "take, produce, use, discard" approach, which results in a significant amount of waste ending up in landfills, thus contributing to the emission of adverse greenhouse gases. [3] Furthermore, there is a large potential in raw materials extracted from the waste. Composting can be considered a natural, organic and renewable method of increasing soil fertility, which does not require any chemical additives. In addition, it has been demonstrated in large-scale projects that by applying compost to agricultural land, significant amounts of $CO_2$ can be captured and stored in the soil in the long term. Composting thus helps to close the loop of the value chain by converting organic waste into fertile soil, which in turn provides the basis for new growth. [4], [5]

## 1.2 Commercial Composting

In Europe, more than 90% of collected biological waste is separated [6]. The corresponding process of composting converts biowaste or organic materials into compost, creating a biochemically stable product that contains microorganisms and is capable of storing carbon [7]. By applying compost to agricultural land, soil properties are improved by increasing organic matter content, PH buffer capacity, and water retention [8]. The most prominent method of commercial composting in central Europe is open windrow composting, in which the organic source material is piled into long, triangular compost windrows. The compost windrows have a base width of 2.5 - 3 m in most industrial plants. The length of the compost windrows may vary considerably, ranging from 35 to 100 m. To enhance the biological process of composting, that is, to stimulate aerobic microbial activity, to release excess heat, to dissipate moisture and to ensure aeration, the compost windrows are mechanically turned. This task of turning the compost windrows is performed by compost turners. Compost turners are tracked machines with a rotating spiked drum designed to turn the compost windrows. By driving the compost turner through the windrows, the rotation of the spiked drum mixes the organic material and piles it up behind the machine. The number of times the compost windrows are turned depends on several factors, including annual rainfall, ambient temperature, and humidity [9]. The process of composting is divided into several phases, with the main rotting phase being the most important stage of the process. Although there are different approaches to how often the compost windrows need to be turned, all have in common that the main rotting requires the highest rate of turning, which decreases in the subsequent phases. One possible approach is to turn the compost windrows every other day in the first two weeks, every third day in the third week, and in the fourth and fifth weeks the compost windrows only need to be turned once a week. [7], [10]

## 1.3 Robotics in Agriculture

The demand for robotics in industrial agriculture and smart farming solutions has increased considerably in recent years. [11] Applications range from robots in the prototype stage, which are mainly used for research purposes, to mature products already deployed in industrial environments. Examples include pruning robots, which are used in the field of tree care. [12], [13] For weed control and fertilization, research is being conducted on spray robots, [14], [15], [16] while other approaches aim to completely avoid the use of herbicides by refining the seedlings [17]. Extensive research is ongoing in the field of harvesting robots, where manual activities such as picking apples [18], sweet peppers [19], [20], [21], [22] or artichokes [23] are being automated. While large-scale deployment of industrial agricultural robots is not yet achieved, ambitious approaches such as Project Xaver [24], Orion [25] and Farmdroid [26] deserve to be mentioned.

Industries with an already high level of automation, such as the automotive industry, are relying on industrial robotic platforms on a large scale, featuring high positioning accuracies and speeds. In agricultural applications, in addition to the requirements for positioning accuracy and speed, the very harsh environment and difficult-to-predict environmental conditions must also be considered. [27]  Since robots in agriculture are often track-driven, it is worth noting that the literature on the modelling and calculation of tracked vehicles can be divided into three categories according to [28]. The first category deals with the analysis of steerability [29], [30], the second category is the analysis of ride characteristics [31], and the third category is theoretical prediction of ground pressure distribution and tractive performance [32], [33], [34], [35], [36], [37]. From a technical perspective, the Robot Operating System (ROS) is of major relevance in the research and development activities, since it enables virtual testing and debugging of processes and algorithms in a simulation-based environment. Research focuses on navigation and control, including motion planning for manipulators [38], simultaneous localization and mapping, and path planning algorithms [39]. In addition to vision-based approaches [40], navigation sensors such as GNSS, LiDAR, and IMU sensors are also used for object detection and path planning. [41], [42]

## 1.4  Aims and Scope

### 1.4.1  Requirements and needs in industrial composting

Increasing automation in the composting industry is imperative. Primarily, the industry recognizes the need to reduce reliance on manual labor due to the lack of skilled labor and the prevalence of high levels of manual labor and adverse working conditions. A fundamental requirement of autonomous machines in the composting industry is that they are capable of performing tasks independently with as little input from an operator as possible. This also applies to the execution of activities at night. Operating a composting facility overnight allows autonomous vehicles to be used more widely, as the risk of collisions with other vehicles is lower due to the lower activity at the composting site. Furthermore, night driving also represents an advantage from the operator's perspective. Many municipalities and cities encourage to move the composting to the night, because strong smelling gases at night are usually less negatively perceived by the population. Furthermore, a requirement arises from the area of quality management, specifically the particularly slow turning of compost windrows. While it is not acceptable from an economic point of view for an operator to drive through the compost windrows particularly slowly, the slow, layer-by-layer turning of the compost results in a particularly air-permeable compost, which ultimately leads to a high quality of the end product.

### 1.4.2  Key Contributions

To meet the challenges industrial composting plants are currently facing, the present work aims to make a contribution to address the requirements for autonomous, self-driving compost turners from an engineering perspective. Therefore, the present work covers the following key contributions:

- We present the development of a self-driving, autonomous compost turner. The main objective of the machine is to perform tasks without operator support at the composting site, primarily the autonomous turning of compost windrows.
- Thereby, we describe aspects of the hardware design, that is, sensor mounting and setup, a navigation module, and an IIoT module for control and data processing tasks. We also present interfaces and interactions of components on a system-wide level.
- Furthermore, we describe the architecture of concepts, models, and software integration. In detail, the navigation tasks consisting of Navigation Filter and Sensor Fusion, compost Windrow Detection algorithm and Route Planning within the compost plant are presented. In addition to the control tasks, we describe the IIoT module's real-time cloud-based processing tasks of compost data." [43]

# 2  Materials and Methods

The present chapter addresses the architecture and structure of the required technical design components to achieve the desired degree of automation of the compost turner. Thereby, both the hardware and the software architecture are considered. The aim of the chapter is to provide in the first step an overview of all the elements necessary for the automation, in order to describe detail aspects in the next step. Thus, the first subchapter is focused on the hardware and sensor setup, while the second subchapter presents the software architecture.

## 2.1  Hardware, Modules and Tools

### 2.1.1  The Electric Compost Turner

Compost turners are used in commercial composting to regularly turn the compost windrows. "For this investigation, a fully electric compost turner, as shown in Figure 1, was used, whereby the machine was developed by an industrial partner. [44] The vehicle is designed for the harsh environmental conditions on a composting plant and therefore runs on two tracks (1). The vehicle travels at a maximum of 0.5 meters per second during the compost turning operation. With a track width of 3.6 meters and a height of 2.2 meters, the compost turner can drive lengthwise through the windrows. A spiked drum (2) is located in the middle of the compost turner. The spiked drum picks up the compost material at the front and ejects it again at the rear of the machine, thereby mixing the compost. The drum has a diameter of 0.55 meters and rotates at a speed of approximately three times per second as it passes through a windrow. Especially during the turning process, that is, when the drum is active, the vehicle is exposed to strong vibrations, which must be taken into account when selecting the navigation sensors. The machine's control cabinet is located underneath the side wall (3)." [43]

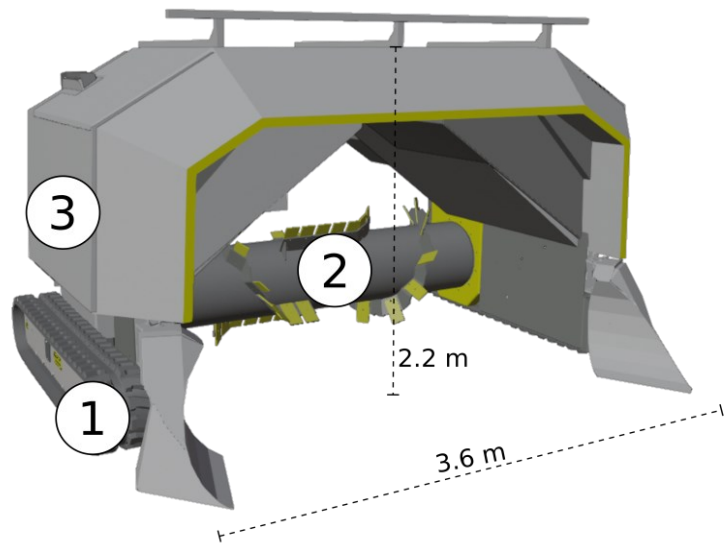

**Figure 1:** The compost turner with tracks (1), drum (2), and location of the control cabinet (3). (own illustration)

### 2.1.2  Navigation Sensors

Especially regarding the aspects of the driverless operation of the compost windrow turning process, a multitude of sensors and actuators were necessary. However, it must be explicitly pointed out that the present work was carried out in the course of a research project with a strong industrial focus. Accordingly, certain sensors, such as the encoders of the drives, were already installed within the compost turner, while other sensors had to be added. Additionally, the compost turner was operating on a daily basis at an industrial composting plant. Therefore, the setup of the sensors, navigation and control modules always had to be performed in such a way, that the daily operations of the compost turner on the composting plant would not be disturbed by the research activities in the slightest. As will be discussed in greater detail subsequently, a strong focus was placed on the design of a modular structure during the entire development phase of the hardware setup. This modular design should allow a quick mounting and dismounting of the sensors to reduce the downtime of the compost turner for the plant operators as much as possible. In this context, a brief note must also be made regarding the design of the control system. As already mentioned, the compost turner was used in ongoing operation apart from the research project. Accordingly, the design of the control system did not only have to satisfy the requirements for modularity. Moreover, a guarantee had to be provided that the research project's control system could be completely switched off at the end of the regular test runs. This reset to the initial state was of utmost importance to ensure the reliable industrial daily operation of the compost turner. Considering the mentioned aspects, the basic structure of the sensors and the automation modules will be presented in the following. Selection of the navigation sensors was an especially significant aspect, therefore this issue was addressed in the course of several series of experiments. [45], [46]

Based on the findings of the previous research activities, as outlined in [10, Ch. 6], the following setup was adopted. In the first step, two dual antenna GNSS receivers with access to network RTK, as well as a stereo camera and a mid-range MEMS IMU were selected. The compost turner used in the experiments had two rotary encoders, which provided the angular velocity of the left and right tracks. In addition, an optical LiDAR sensor was added to this setup. Table 1 shows a summary of the used sensors.

| Type of Sensor | Quantity | Model | Description |
|---|---|---|---|
| **GNSS** | 1 receiver 2 antennas | Alberding A12-RTK | geodetic GNSS receivers |
| **IMU** | 1 | XSens MTi-G-710 | MEMS IMU |
| **Odometry** | 2 | Atech AC-X | Wheel encoder on Compost Turner's tracks |
| **Optical Sensor** | 1 | Velodyne Ultra Puck | LIDAR |

**Table 1:** Employed sensors

The aforementioned modularity of the sensors was ensured by a mounting frame. The basic idea was to provide an option for quickly and precisely attaching the sensors (and subsequently detaching them again), while minimizing the need to modify the outer surface of the compost turner. In consultation with the industry partner, a two-part aluminum structure was created. The first part consisted of three L-shaped rails, which were attached to the compost turner using screw connections, see Figure 2. This part was thus permanently attached to the compost turner and was not removed during daily operation on the composting plant.

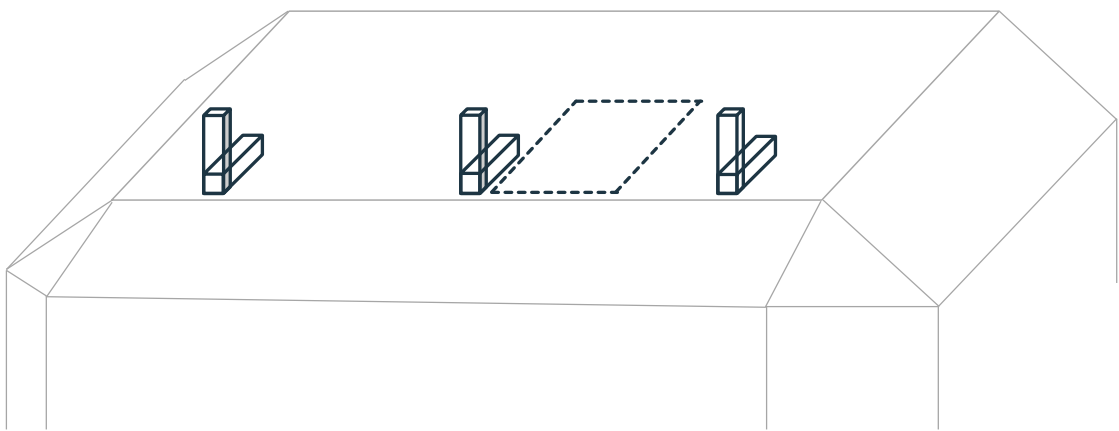

**Figure 2:** The L-shaped rails were permanently attached to the compost turner and form the basic framework for the sensor rail. The dashed line shows the area where the navigation module, which contains the processing electronics, is attached.

The second part consisted of an aluminum profile to which the sensors were attached. This part was not fixed and could therefore be quickly and reliably attached to the L-rails. As shown in Figure 3, the two GNSS antennas, the IMU, the stereo camera and the LiDAR were attached to the profile. The inclination of the LiDAR could be adapted via an adjustment platform. This was significant since a purely horizontal mounting of the LiDAR would not provide the possibility to detect objects standing directly in front of the compost turner. By tilting, the field of view of the LiDAR could be adjusted accordingly. A more detailed discussion of the determination of the optimal tilt angle is given in the following chapter.

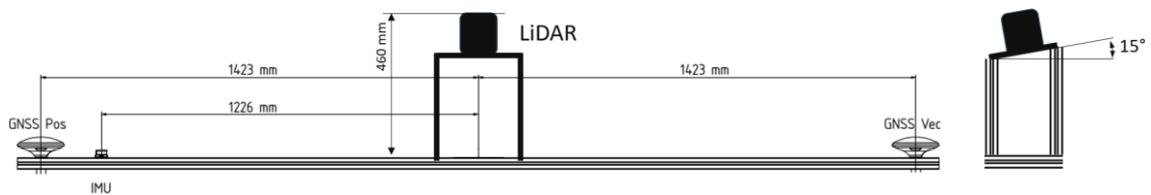

**Figure 3:** The sensors were attached to an aluminum profile, which could be quickly and reliably attached to the L-rails on the compost turner (Figure 2). The degree of inclination of the LiDAR sensor can be variably adjusted as required.

### 2.1.3  The Navigation Module

For the processing of the sensor data and the subsequent computation of the navigation tasks, a module was designed containing the required evaluation electronics and computational capabilities. As with the sensor rail already presented, it was required that the navigation module could be quickly mounted and dismounted on the compost turner. A schematic representation of the module is given in Figure 4. The module consists of an Embedded Computing Board (ECB), a CAN board, GNSS receiver, power supply and a battery. The Robot Operating System (ROS) version Noetic is installed on the ECB. A detailed execution of the workflows in ROS will take place in the following chapter. Since previous research projects found that an Nvidia Jetson Nano has insufficient computing capacity, an Nvidia Jetson Orin was chosen as the ECB.

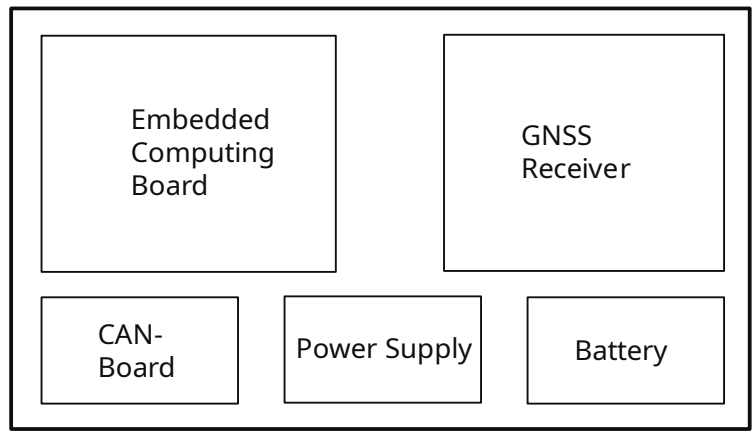

**Figure 4:** Schematic representation of the navigation module, consisting of the components embedded computing board (ECB), GNSS receiver, CAN board, power supply and battery.

### 2.1.4 IIoT-Module for Control Tasks and Cloud-based Data Visualization

Figure 4 provides an illustration of the IIoT module, outlining the hardware components used: programmable logic controller (PLC) with CAN bus adapter, an industrial LTE router, a switch and a DC-DC voltage converter. The two dust-tight RJ45 connections provide LAN connectivity, the two LTE antennas form the basis for the wireless connection to a web server, and the Wi-Fi antenna is used for local control of the IIoT module.

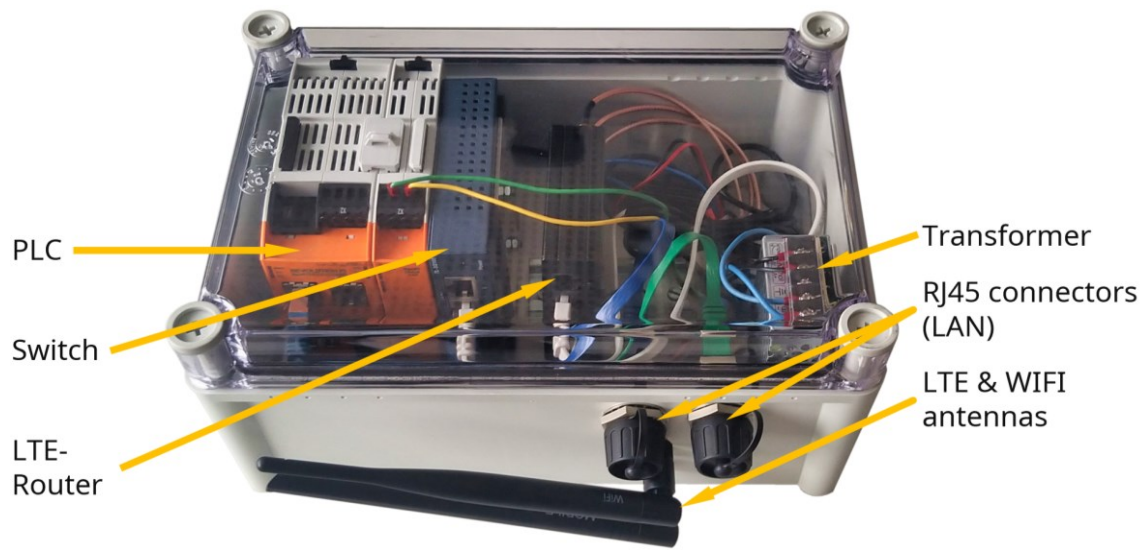

**Figure 5:** Illustration of the assembled IIoT module. The module is responsible for the compost turner's control tasks and for transmitting sensor and navigation data to a web server.

## 2.2 System Architecture – Hardware Integration in the industrial Compost Turner

After the sensors, navigation and IIoT modules were presented in detail in the previous chapter, the aim is now to explore on a system-wide level how these components were integrated into the overall architecture of the industrial compost turner. Figure 6 provides a schematic representation of the relevant components, focusing on the hardware. A more in-depth discussion of the software architecture is given in the subsequent chapter.

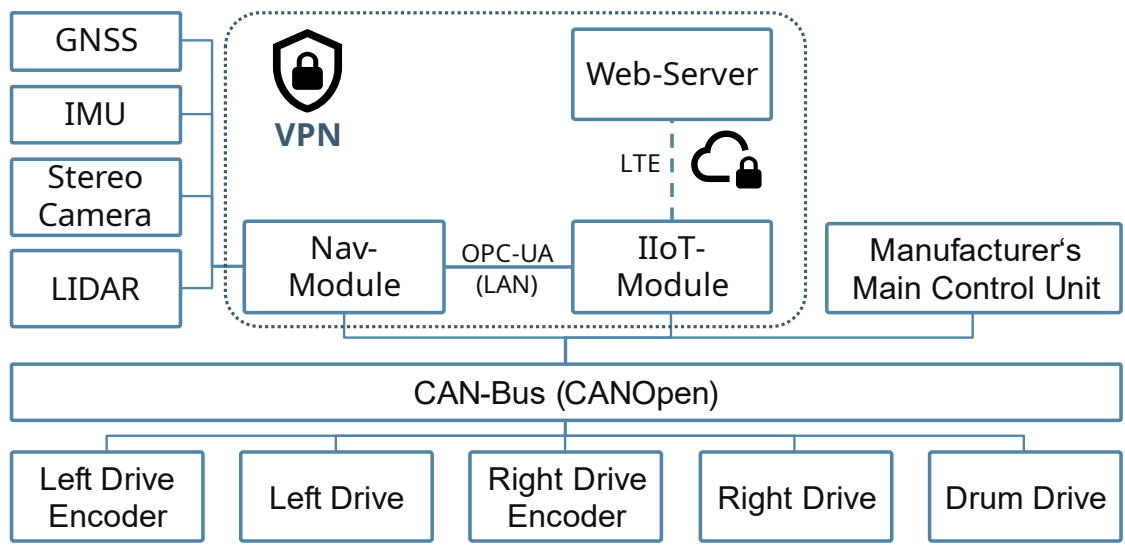

**Figure 6:** High-Level view of Modules, Sensors, and Actuators within the system autonomous Compost Turner

The integration of modules and sensors into the existing system of the industrial compost turner is illustrated in Figure 6. The drives of the tracks (Left Drive and Right Drive), their encoders (Right Drive Encoder, Left Drive Encoder), and the drive of the drum (Drum Drive) are permanently integrated in the compost turner and send regular sensor data and status logs to the CAN bus. However, only the Manufacturer's Main Control Unit is directly allowed to control the drives. Even if it were possible from a technical point of view to control the drives directly via the CAN bus, for example from the IIoT module, this would involve considerable safety risks. Due to the ongoing operation of the compost turner, it can be assumed that regular maintenance work will take place on the hardware and software. An address conflict on the CAN bus between the manufacturer/maintenance team and the research team must therefore be avoided at all costs. To circumvent this problem, the drive commands were sent in the first place from IIoT Module to the Manufacturer's Main Control Unit, and subsequently from there to the drives. In this way, the drives only receive the drive commands from the Manufacturer's Main Control Unit. The described setup is shown in Figure 6. All the sensors already presented are connected to the navigation module, which can also read the status and sensor data from the CAN bus. The IIoT module communicates with the manufacturer's main control unit via the CAN bus and with the navigation module via OPC-UA. Communication with the navigation module via the CAN bus would be technically possible, but is less flexible than via the OPC-UA protocol, as will be explained in more detail. The sensors used are connected via manufacturer-specific connections to the navigation module, which also provides their power supply. The navigation module, like the IIoT module, is connected to the internal 24V power supply of the compost turner. The web server is also reached via the OPC-UA protocol, with connectivity provided by the IIoT module's LTE capability mentioned in the previous section. Due to the fact that the compost turner operates in an industrial environment, data security is of utmost importance. Although the OPC-UA protocol is encrypted by default, an additional layer of security was created by establishing a virtual private network (VPN). At this point, it should be noted that it was not only the sensor data, could be accessed via the LTE connection. The entire compost turner could be controlled via the LTE capability of the IIoT Module.

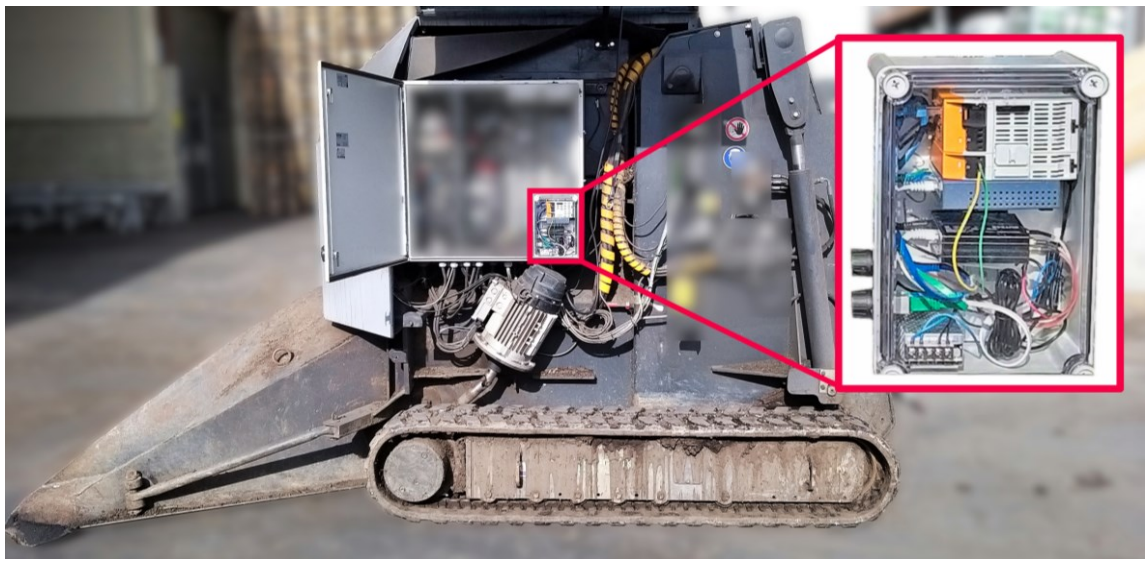

**Figure 7:** Integration of the IIoT module into the circuit box of the autonomous compost turner. The module is designed for plug & play use: once the cables are connected, all protocols automatically boot up and subsequently enter operational mode.

## 2.3  Architecture of Concepts, Models and Software Integration

The Robot Operating System (ROS) forms the foundation of the software architecture. Of particular advantage was that ROS offers the possibility to assign tasks to the different sensors used within a common framework. Thus, a large part of the software development could be done exclusively in ROS. Two main subject areas can be defined, as shown in the high-level view in Figure 8. On the one hand, there is the subject area of evaluation and processing the entire navigation sensors signals in order to calculate an optimal trajectory for the compost turner. On the other hand, the second main subject area is the control of the tracks, which receives the calculated target routes as input and sends the control commands to the drives as output. Both topics will be presented in more detail within the following chapters.

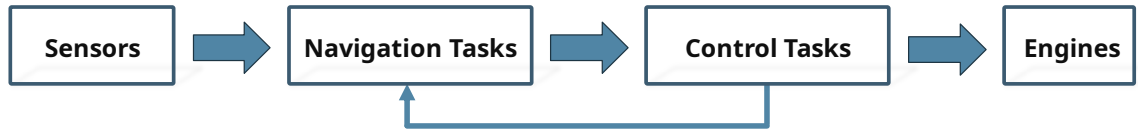

**Figure 8:** High-Level view of the Software Architecture. The two main tasks are Navigation and Control

### 2.3.1 High-Level View on Navigation Tasks

In the software implementation of the navigation tasks, the publish/subscribe communication between nodes integrated in ROS was a major advantage. This functionality enables several nodes to subscribe to a topic at the same time and to process the received data simultaneously. Due to the sufficient computing power provided by the corresponding hardware in the navigation module, data processing, visualization and data storage could be performed simultaneously. An overview of the entire program flow in ROS is shown in Figure 9. This is an extended version of the approach presented in [10]. As can be seen in the figure, the sensors, shown in red, provide data that is collected via the sensor nodes shown in blue. In these nodes, the data is packaged into a ROS message, and then published to the ROS network. Via the ROS master, all other nodes can subscribe to the sensor nodes. In this context, the nodes for recording, visualization and the node of the Extended Kalman Filter (EKF), which will be discussed in detail, are of particular importance. Another advantage in ROS is that the entire recorded data can be saved in a .bag file format. These ROS bag files can be played back in post-processing to simulate the conditions of a real-world scenario. This technique facilitated the simulation-based development of the ROS architecture.

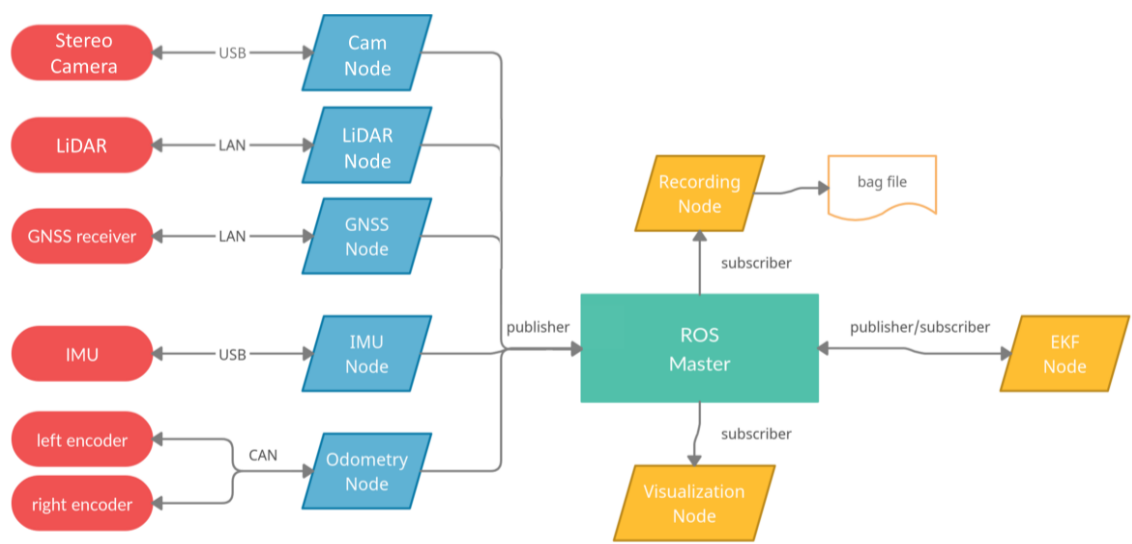

**Figure 9:** ROS Program Flow for Navigation Tasks. Sensors are displayed as red oval shapes, Sensor Nodes in blue, the ROS Master is displayed in green, and the subscriber nodes in yellow. Recording of the data is done via the bag file.

### 2.3.2   Navigation Filter

The fusion of sensor data represents an essential aspect for the calculation of optimal routes for the compost turner. Due to the real-time requirements, an approach with the most efficient computing power was sought, leading to the choice of the Error-State Extended Kalman Filter (EKF), as discussed in [10, Ch. 4.3]. Specifically, two different approaches were compared, a modified federated integration architecture and a cascaded integration architecture. [10, Ch. 5.3]

### 2.3.3   Windrow Detection

To accurately and efficiently maneuver a compost turner through the compost windrows of a composting facility, a detailed global point cloud of the facility is created using a LiDAR scanner mounted at a specific tilt on the compost turner. To create this cloud, the scanner is manually moved around the compost windrows. The LiDAR point cloud is then georeferenced to create a comprehensive global point cloud, and insignificant data is filtered out to reduce computational effort. Individual compost windrow clusters are then extracted from the generated global point cloud. A RANSAC procedure is used to remove outliers that could affect the overall representation of the clusters. Euclidean clustering is then used to identify and isolate the individual rows. After the windrow clusters are classified, they are analyzed to find the optimal route for the compost turner through each windrow. The optimal route should follow the edge of each row, which requires extracting edge points from each row cluster. Through various steps involving additional RANSAC filtering and calculations of point coordinates, a set of points representing the ridge of the windrows is obtained. This data is used to find the start and end points of a line segment that represents the optimal route through the compost windrow.  [43], [47]

### 2.3.4   Route Planning

The routing algorithm for the compost turner's global path within a composting plant is determined once the most efficient path through each windrow has been identified. The strategy aims to determine a path that includes selected windrows, taking into account the space requirements of turning maneuvers by the compost turner within the confined space available at the composting facility. This may include moving from the end of one windrow to the beginning of another, or moving to a particular destination with a specified destination, such as a loading station. The implementation of the algorithm provides users with the ability to freely choose the preferred windrows to be processed by the autonomous compost turner. The entire route is a set of waypoints, with each windrow symbolized by two unique waypoints - the start and end points of the windrows. [43], [47]

### 2.3.5   Control Tasks of the IIoT Module

The IIoT module was methodically designed using an Model-based Systems Engineering Approach and includes two sub-modules. The former is responsible for control tasks of the compost turner's tracks and will be discussed within the present section. The following section will deal with the latter subsystem, which is responsible for acquisition, processing, transmitting to web server and visualization of compost data.

The ROS framework offers numerous packages which can be freely accessed and adapted. Besides the obvious advantages of open-source packages, however, increased attention must be paid to the reliability of the software implementation. Due to the industrial environment in which the compost turner operates, a robust control of the tracks was of major importance. The mentioned robustness does not only refer to the theoretical control error, that is the deviation from the target trajectory. In particular, the software implementation had to meet industrial standards, since due to the high drive power of the compost turner, a false control command would have severe consequences.

Various approaches have been pursued in the literature to design control strategies for tracked vehicles. Starting from standard methods using PID controllers, there are increasingly promising approaches in the area of model predictive control (MPC) for tracked vehicles. [48] This MPC method tries to minimize the tracking error by applying a dynamic vehicle model whose future driving behavior is estimated within a given prediction horizon. For example, [49] presents an MPC motor torque control system based on the combination of MPC and Extended Kalman Filter (EKF). The problem encountered, however, was that the initial version of the control system was too slow to respond to the rapidly changing load distribution of the vehicle. Therefore, the kinematic model used to determine the future state of motion was simplified by not including slip in the model. The torque of the drives was set as the control variables of the MPC in order to obtain a higher dynamic response of the control. The aim of MPC was to minimize the cost index, which is composed of lateral, heading and speed deviation of the trajectory. Other research sees potential in the use of MPC for tracked vehicles by making an improved distribution of the weight coefficients of the objective function. Specifically, in a set of experiments different weight coefficients of the objective function were tested in a kinematic model of the tracked vehicle to determine the relationship between coefficients and tracking error. [50] Other notable approaches to the use of MPC for tracked vehicles is [51], where the aim is to provide the most efficient energy consumption of the vehicle. In the agricultural field, [52] developed an MPC for a tracked robot, which was designed for harvesting lettuce, although an industrial application scenario was not mentioned.

### 2.3.5.1   Inital Approach: A Model-Predictive-Control for Tacked Vehicles
Due to the promising capabilities of the MPC in the field of tracked vehicles, this method was also employed for the compost turner. Even if this approach did not finally succeed, the considerations made will nevertheless be briefly presented.

Based on an internal dynamic model, the basic approach of model predictive control is to minimize the cost function $J$ over the receding horizon $N$. This is accomplished by an optimization algorithm that minimizes the cost function $J$ by varying the control input $u$. Thereby, the choice of specific solver can be adjusted depending on the intended usage. [53] The quadratic cost function $J$ can be written as

$$J = \sum_{k=0}^{N} \left( \mathbf{x}(k)^T Q \mathbf{x}(k) + \mathbf{u}(k)^T R \mathbf{u}(k) \right)$$

where the vector $\mathbf{x}$ is composed of the difference between the actual position of the vehicle $[x, y, \varphi]^T$, and the desired position $[x_{set}, y_{set}, \varphi_{set}]^T$

$$\mathbf{x} = \begin{bmatrix} x \\ y \\ \theta \end{bmatrix} - \begin{bmatrix} x_{set} \\ y_{set} \\ \theta_{set} \end{bmatrix}$$

The control input vector $\mathbf{u}$ consists of the longitudinal throttle input $v$ and the steering input $\omega$, with constraints on maximum acceleration and maximum velocity imposed on both quantities.

$$\mathbf{u} = \begin{bmatrix} v \\ \omega \end{bmatrix}$$

The matrices $Q$ and $R$ weight the state vector and control input vector. The structure of the cost function $J$ is shown again in Figure 10. As can be seen, the first part of the cost function is responsible for path tracking. The aim is to minimize the deviation $\Delta$ from actual position to target position in $x$ and $y$ direction, as well as the deviation of the orientation by the angle $\varphi$. The weighting matrix $Q$ in this case is a diagonal matrix whose entries are a measure of the priority of minimizing the individual deviations $[\Delta x, \Delta y, \Delta \varphi]$. The second part of the cost function $J$ refers to weighting of the control vector, that is, the longitudinal throttle commands $v$ and the steering angle $\omega$. The larger the values of the diagonal matrix $R$ are, the more the respective driving command is penalized. Too low entries for $R$ thus lead to very fast changing, jerky driving commands. On the other hand, too high entries in $R$ cause long and sluggish travel commands, whereby the trajectory can no longer be followed.

$$J = \underbrace{\sum_{k=0}^{N} \mathbf{x}(k)^T Q \mathbf{x}(k)}_{} + \underbrace{\sum_{k=0}^{N} \mathbf{u}(k)^T R \mathbf{u}(k)}_{}$$

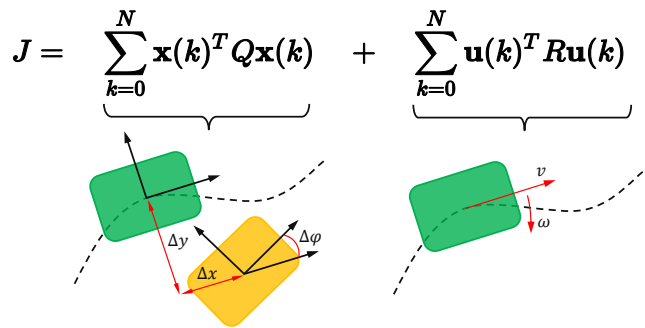

**Figure 10:** The first part of the cost function $J$ is responsible for path tracking and is set via the weighting matrix $Q$. The second part is responsible for the control input, consisting of linear and angular components, and is penalized via the weighting function $R$.

As often found in the literature, the kinematic model of a tracked vehicle with the neglection of slip occurrence was used in the first step. Based on the equations of motion for differential drive robots, the simplified kinematic model of a tracked vehicle in world coordinates can be written as: [54]

$$\begin{bmatrix} \dot{x} \\ \dot{y} \\ \dot{\varphi} \end{bmatrix} = \begin{bmatrix} \dfrac{r}{2}\cos(\varphi) & \dfrac{r}{2}\cos(\varphi) \\ \dfrac{r}{2}\sin(\varphi) & \dfrac{r}{2}\sin(\varphi) \\ -\dfrac{r}{b} & \dfrac{r}{b} \end{bmatrix} \begin{bmatrix} \omega_{L} \\ \omega_{R} \end{bmatrix}$$

Thereby, $[x, y, \varphi]^{T}$ represent the coordinates and orientation of the vehicle. Further given are the width of the vehicle $b$ and the radius of the sprockets $r$. The angular velocities of the left and right sprocket are represented by $\omega_{L}$ and $\omega_{R}$.

The implementation of the presented Model Predictive Control approach was performed in Python based on the open source toolbox do-mpc.[55] As shown in Figure 11, this approach yields promising results in the simulation. The target trajectory shown in red can be well followed by the actual trajectory shown in blue.

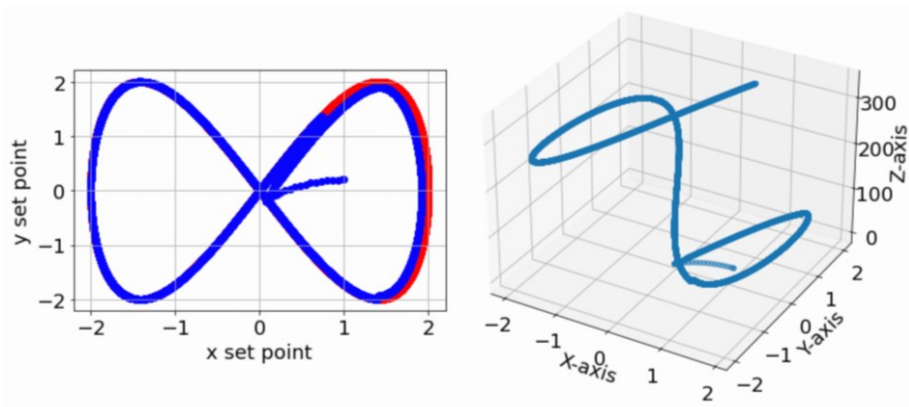

**Figure 11:** Initial simulation results of the Model Predictive Control: The target trajectory shown in red can be well followed by the actual trajectory shown in blue.

However, the technical integration of the MPC in the industrial environment of the autonomous compost turner proved to be particularly challenging. Even though the initial simulation results of the Model Predictive Control approach yield promising results, the software-technical implementation could not achieve a level of robustness that would allow its use in the real compost turner. To circumvent this barrier, an approach within the ROS framework was pursued further, as presented in the subsequent chapter.

### 2.3.5.2   Industrial-proven Differential Drive Control leveraging ROS

ROS-industrial is an international open-source project that promotes the use of ROS in the industrial environment of robotics and manufacturing automation. One of the ROS packages for control applications funded in this framework is ros_control. [56] The package provides predefined interfaces for controller, controller manager and hardware interfaces. Due to its broad applicability, the package is used in many areas of research and industry.

A sub-package of ros_control provides a differential drive controller. In contrast to the model predictive control, the differential drive controller is less complex, but offers the advantage that only algebraic equations must be solved instead of differential equations. This leads to a significantly lower computing time, which in turn contributes to the stability of the entire ROS framework. At this point, it should be noted that the differential drive controller is subject to significantly less theoretical potential in terms of control theory compared to the MPC. As already mentioned, the implementation of the open-source package ros_control, which has already been tested in an industrial environment, as well as the existing integration in the ROS framework were decisive for the choice of the differential drive controller.

From a mathematical perspective, the differential drive controller can be easily formulated by deriving the inverse kinematics of the differential drive robot's equation of motion. As shown in Figure 12, $V$ is the linear velocity in the direction of travel, denoted as `lin_vel` in ROS. The steering input $\omega$, is denoted as `ang_vel` in ROS. Furthermore, the width of the vehicle $b$ and the radius of the sprockets $r$ are given. In the specific case, the radius corresponds to the radius of the sprockets of the compost turner.

Starting point for the robot's equation of motion is the equation $R = V/\omega$, and the relation of the angular velocity to the linear velocity for both sides. That is, for the right sprocket $\omega_R = v_R/r$ and for the left sprocket $\omega_L = v_L/r$. Substituting both above mentioned equations in the definition of the angular velocity,

$$\omega \cdot (R + b/2) = v_R$$
$$\omega \cdot (R - b/2) = v_L$$

where $R$ is the distance from the center of the vehicle to the instantaneous center of rotation ($ICR$), one obtains the angular velocities of the right wheel $\omega_R$ and left wheel $\omega_L$

$$\omega_R = \frac{V + \omega \cdot b/2}{r}, \qquad \omega_L = \frac{V - \omega \cdot b/2}{r}$$

Thereby, the angular velocities of the wheels are, in the present case, the angular velocity of the sprockets of the compost turner.

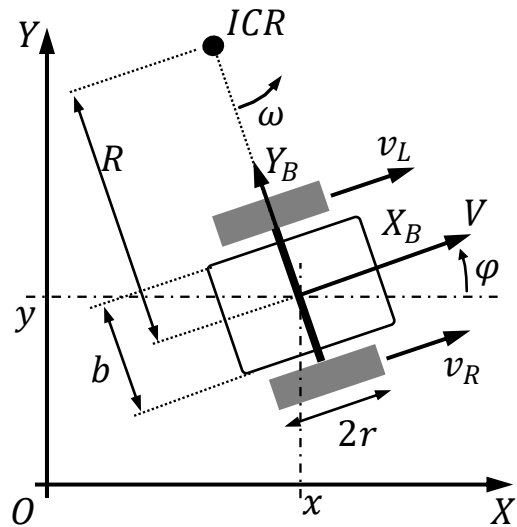

**Figure 12:** schematic representation of a differential drive robot

The Diff Drive Controller's input, that is, the linear and angular velocity, are constrained with respect to maximum the velocity, as well as minimum and maximum acceleration. This is significant because, even if the controller misbehaves, as for instance, a discontinuous jump as control input, this drive command is first smoothed over the constraints before it is passed on to the real machine.

A high-level overview of the control workflow is shown in Figure 13. The starting point is the command velocity calculated in the navigation task, consisting of the linear component $V$ and the angular component $\omega$. The diff_drive_controller is loaded into a controller manager and a configuration of the parameter set is performed in advance, such as defining the velocity and acceleration constraints. The control output is obtained by accessing the ROS hardware_interface. It is worth noting that the underlying idea of the hardware_interface would be to send the control output directly to the drives of the robot. However, due to the demands on robustness, a different approach was chosen for this project. Specifically, the control outputs were published to a communication node, which forwards them via Ethernet to a programmable logic controller (PLC) using the OPC-UA protocol. The PLC receives the transmitted data and performs a further error analysis. The limits already set with regard to the maximum speed of the drives are verified again and limited in the event of non-compliance. The hardware and software environment of the PLC, which is considered to be particularly reliable, can thus catch potential errors stemming from the Robot Operating System. In the final step, the commands are sent to the compost turner's manufacturer controller, which forwards the commands to the drives. This last step in the workflow was necessary as only the compost turner's manufacturer has direct access to the drives.

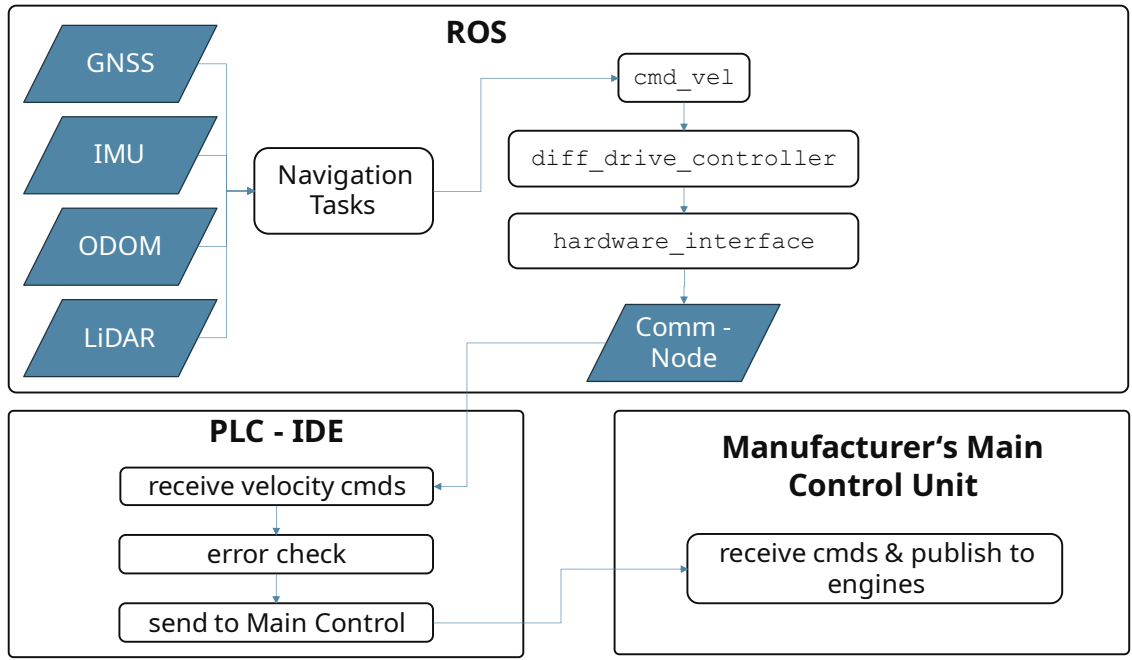

**Figure 13:** Software setup in ROS, PLC, and the Manufacturer Main Control Unit.

### 2.3.6   Cloud-based Data processing tasks of the IIoT Module

The underlying idea of the cloud-based data processing task of the IIoT module is that composting data should be made available to the plant operator in real time. Of particular significance are concentrations of carbon dioxide ($CO_2$) and methane ($CH_4$), as well as the core temperature of the compost windrows. These data shall be automatically processed and prepared, and then transferred to a web server. At the web server a post-processing is performed, the data are stored in a database and visualized on a web frontend. These tasks were methodically addressed in a model-based systems engineering approach. Therefore, a high-level workflow is now presented, as Figure 14 displays. Starting from the sensors, which are connected to the CAN bus via the CANopen protocol, the data is sent to the navigation module in the first step. At this point, a initial pre-processing of the data takes place. The idea is to assign a geo-reference (geotag) and a time stamp (time tag) to the gas and temperature data. Thus, a visualization of all windrows at the compost plant can be displayed afterwards on the web interface, and methane content and temperature can be displayed. Utilizing the OPC-UA connection to the IIoT module, the data is forwarded after the pre-processing step. In the IIoT module, the data is compressed and sent to a web server using the module's LTE capability. The transmission takes place encrypted via the VPN. At the web server, the compost data is stored in a database and displayed visually in a frontend after post-processing. Regarding the visualization, it should be noted that the operator requires a visualization of the compost data over a period of several months, while daily fluctuations do not contribute significant information concerning the quality of the compost.

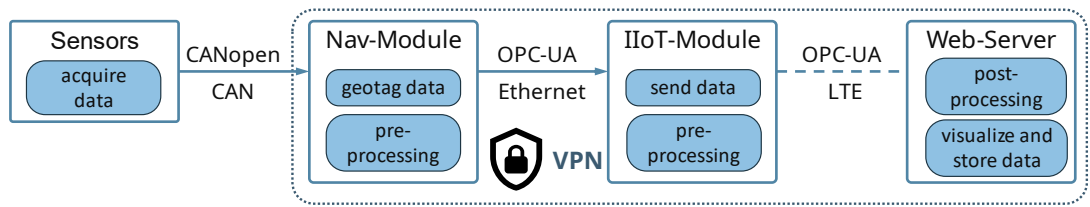

**Figure 14:** A high-level overview of compost data processing

# 3  Results

The preceding methods chapter covered the design and architecture of aspects of the hardware, starting from the selection of sensors up to the high-level view of sensors, navigation, and control module. Equally, the employed mathematical methods, models and algorithms and the resulting system architectures have been presented. The aim of the present section is to demonstrate that the resulting overall system of the autonomous compost turner meets the required expectations as outlined in the introduction chapter. To demonstrate the capability of the autonomous compost turner, three clearly defined experimental scenarios will be presented. The setup of the scenarios is strongly based on standard procedures as commonly found in the respective scientific literature. In the first scenario, the ability of the compost turner to follow a curve with a predefined radius as accurately as possible is examined. The specified curve is therefore a circle, and a comparison is performed between the target trajectory and the actual trajectory. [57] The second scenario is an extension of the first one. The driving command of the vehicle under test is a continuously changing steering input, and the predefined target trajectory is a Lissajous curve in the form of an infinity shape. [58] To be consistent with domain-specific literature, an application-specific capability of the autonomous vehicle will be demonstrated in the third scenario. [59], [60], [61] In the specific case, driving towards a windrow, precision adjustment and alignment in front of the windrow, and subsequent turning of the compost windrow, will be demonstrated. All scenarios were carried out at an industrial composting plant.

## 3.1  Scenario 1: Validation of Waypoint Navigation

In the first test scenario, the given trajectory is a circle, which is traversed by the compost turner with left steering angle. Due to the simple shape of the trajectory in the form of a circle, a validation of the waypoint navigation can be easily carried out. The results are shown in Figure 15 and Figure 16. As clearly evident in the illustrations, there are only minor deviations between the target trajectory and the actual trajectory. The compost turner can therefore follow the target trajectory without difficulty. As can be seen from the velocity curves of the tracks in Figure 16, the autonomous compost turner always heads for a target point on the circle. As soon as it reaches this point, it moves on to the next target point. This waypoint navigation was deliberately designed in this way, as it offers advantages for navigating within the narrow situation of the composting plant. For the specific case of a circular trajectory, there is an apparent oscillatory behavior. In practice, however, this oscillating behavior is hardly noticeable. As can be seen from Figure 16, the 2D error, which represents the difference between the desired target trajectory, and the actual trajectory in the xy-plane, is always below 0.02m. Thus, it is clearly noticeable in practice that the target trajectory in Figure 15 can be followed very well, with only slight deviations.

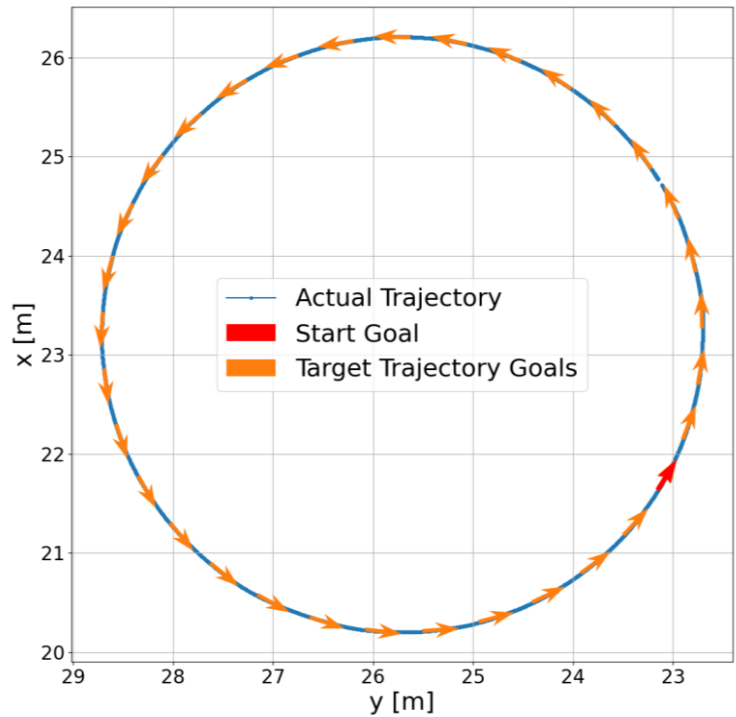

**Figure 15:** Scenario 1. Representation of the target trajectory versus the actual trajectory in a 2D plot.

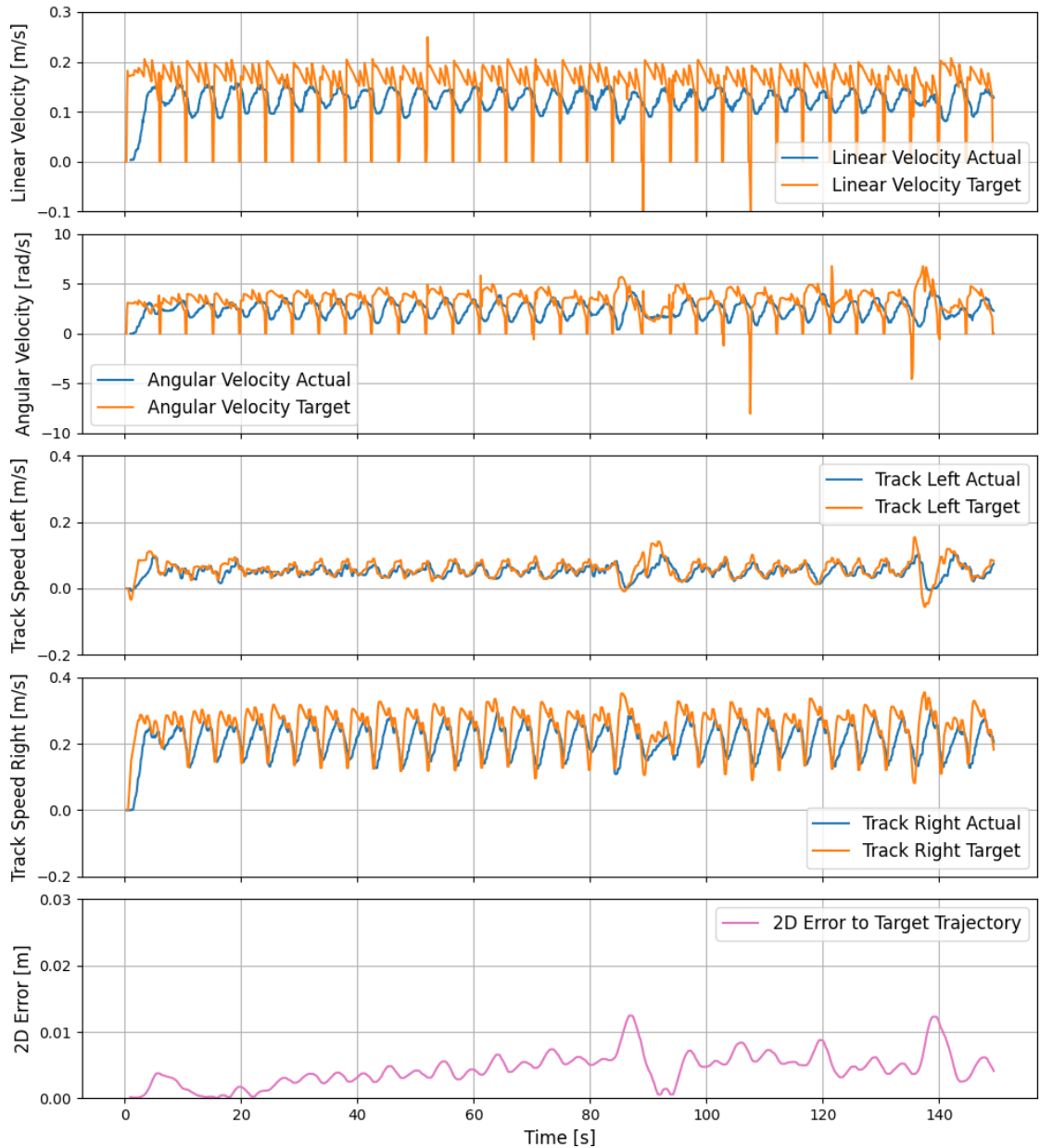

**Figure 16:** Scenario 1: Actual (blue) versus target (orange) velocities of the robot and tracks, and 2D error between actual and target trajectory.

## 3.2 Scenario 2: Bernoulli Lemniscate - Infinity Shape

The given trajectory in the second scenario was a Bernoulli Lemniscate which is defined as follows and results in an infinity-shaped curve.

$$x = \frac{a \cdot \cos(t)}{1 + \sin^2(t)} \; ; \quad y = \frac{a \cdot \sin(t) \cdot \cos(t)}{1 + \sin^2(t)}$$

Where $a$ is the radius and the angle $t$ runs from 0 to $2\pi$. Figure 17 presents the results of the second scenario. As in the first scenario, the Figure 17 shows the actual trajectory and the target trajectory in an X-Y plot, where Figure 18 shows the actual and target trajectories of linear velocity, angular velocity, track speeds, and the 2D error. In scenario 2, the target trajectory, the Infinity Shape, was also divided into discrete target points, to which the compost turner drove in sequence. As a result, the apparent oscillatory behavior can also be seen in Figure 18, which, however, only results from the successive approach of the target points. As can be seen in Figure 17, scenario 2 also demonstrates that the trajectory of the infinity shape can be well followed without any major deviations.

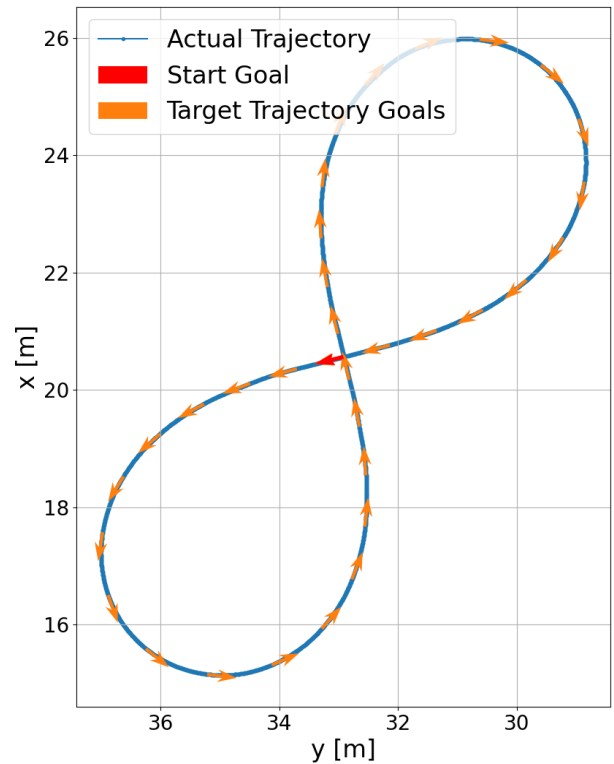

**Figure 17:** Scenario 2. Representation of the target trajectory versus the actual trajectory in a 2D plot.

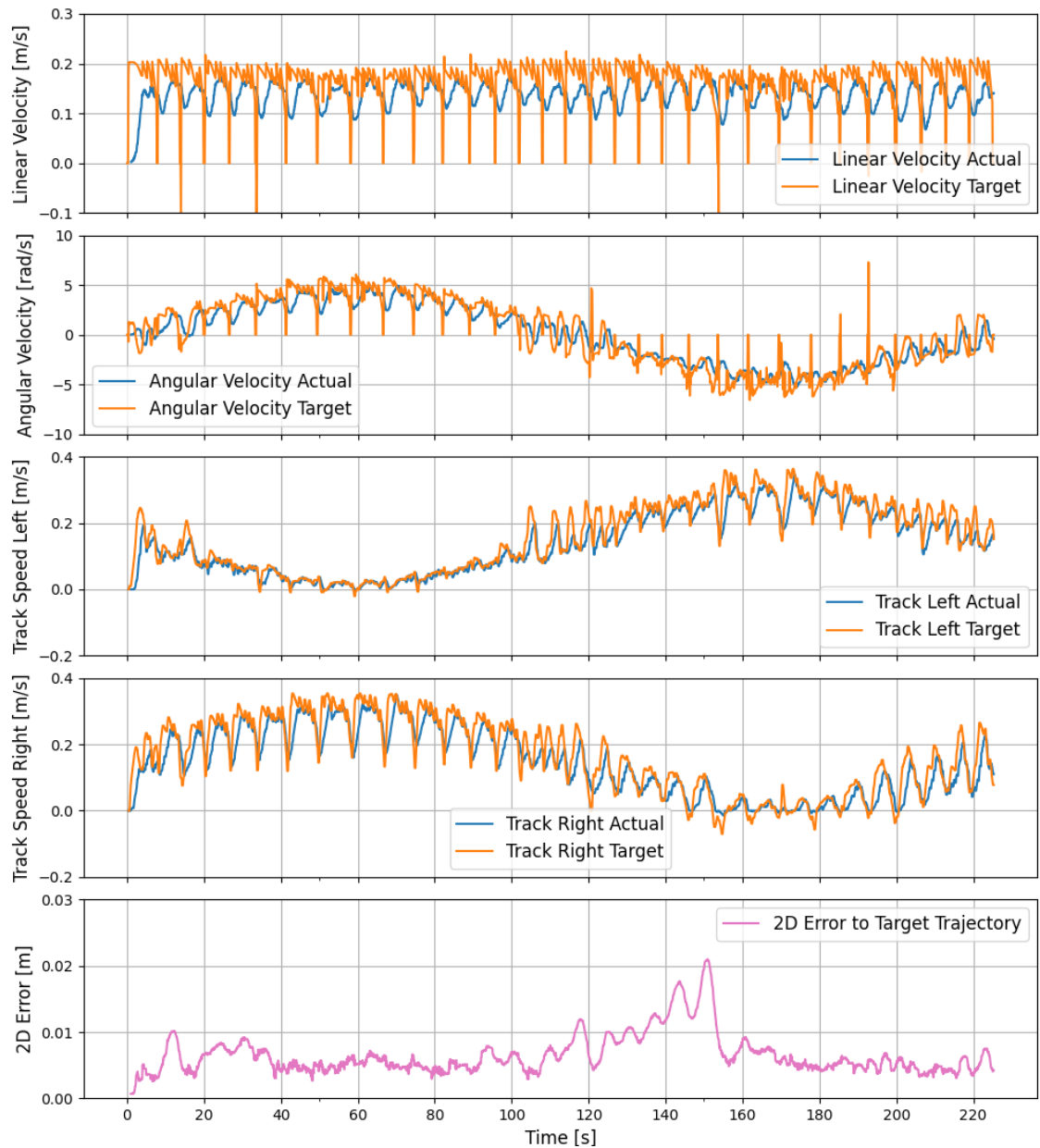

**Figure 18:** Scenario 2. Actual (blue) versus target (orange) velocities of the robot and tracks and 2D error between actual and target trajectory.

## 3.3  Scenario 3: Turning a Windrow

Scenario 3 demonstrates the main task of the compost turner, which is the autonomous turning of compost windrows. The task of the compost turner is to turn the compost windrow as straight as possible. However, the challenge is that the compost windrows are piled up with wheel loaders, which means that they are not always perfectly straight. The compost turner must therefore work against this deviation to produce a straight compost windrow. Scenario 3 will now demonstrate the turning process. As shown in Figure 19, the compost turner is at its assigned position at the composting plant. When the control command to turn a specific compost windrow is issued by the operator, the compost turner moves towards the assigned windrow. Clearly shown in Figure 20 is that the compost turner drives towards the compost windrow on the road and then performs a positioning maneuver to be aligned with the compost windrow. Figure 21 shows that the compost turner's drum is started once it has reached the compost windrow. During the turning process, the compost turner attempts to follow the navigation targets in the compost windrow and sends respective heading adjustment commands to the left and right tracks, as can be seen in Figure 21. Once the end of the compost windrow is reached, the drum velocity is ramped down. The compost turner can now move to another compost windrow, if this was requested by the operator. In this scenario, the given task was turning only one compost windrow, however, turning further windrows of the composting plant in the same run, would certainly be possible from a technical perspective.

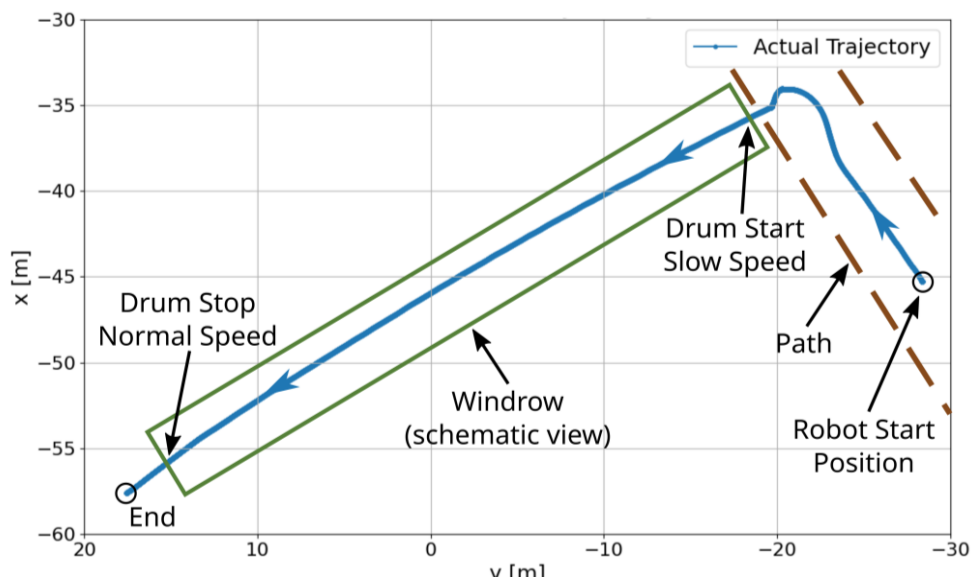

**Figure 19:** Scenario 3. Representation of the actual trajectory in a 2D plot. In addition, the compost windrow is shown schematically.

In terms of the IIoT module and its data transfer capabilities, it can be concluded that it performed effectively in all three scenarios. As will be explained in the discussion chapter, there were dependencies on industry partners, which led to the non-integration of the temperature sensors in the compost turner. Consequently, it was not possible to validate the IIoT module with temperature data, however, this did not pose an issue from a technical perspective. All the data shown in the diagrams of Figure 14 was successfully transmitted to the laptop base station using the LTE capability of the IIoT module. Given that there is no technical distinction between the transmission of temperature data and driving data via the IIoT module, it remains appropriate to consider the module as validated for its designed purposes.

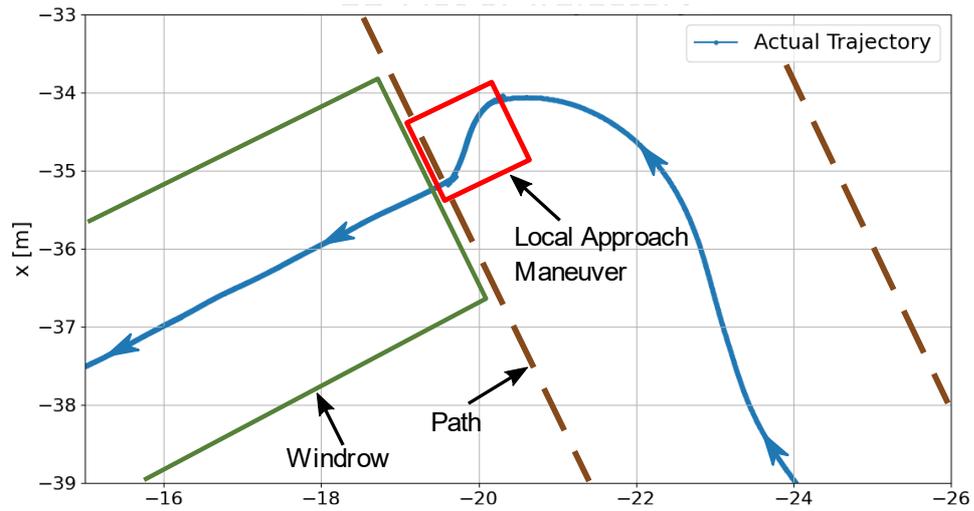

**Figure 20**: Detailed view of Figure 19. Upon reaching the designated compost windrow, the compost turner initiates a repositioning maneuver, aiming to achieve an optimal alignment with the compost windrow.

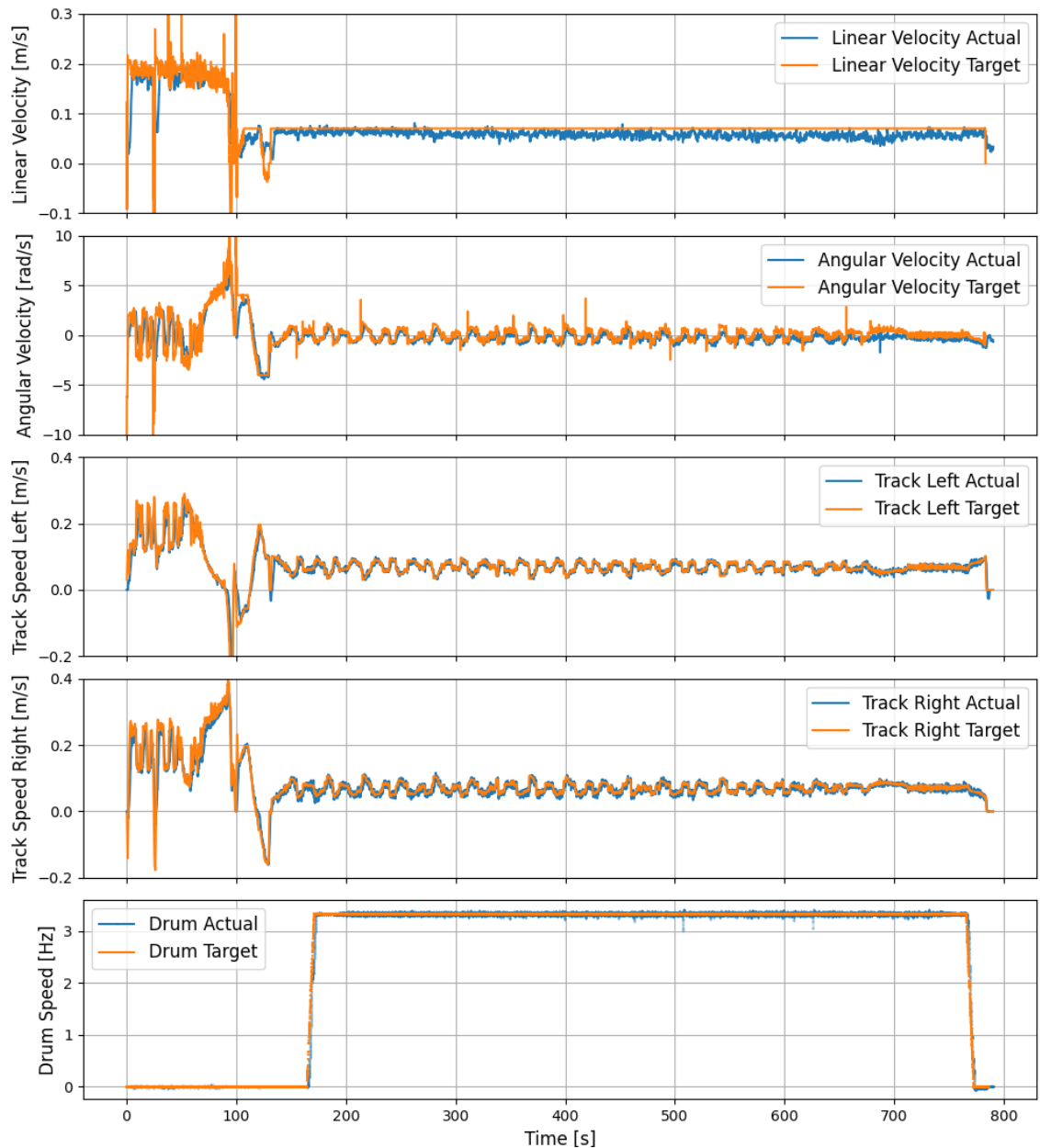

**Figure 21:** Scenario 3. Actual (blue) versus target (orange) velocities of the robot, tracks, and drum.

# 4  Discussion

Thus, research results and development steps for the realization of an autonomously driving compost turner were presented. On the one hand, aspects of the hardware, starting from the sensor selection to the design of modules for navigation and control tasks, were introduced. Furthermore, the required models, algorithms, and system architectures as well as their subsequent technical implementation were demonstrated. In particular, the industrial proven ros_control package for controlling the compost turner, were discussed. Figure 22 summarizes the developed main functionalities of the autonomous compost turner. On the left side, the selected navigation sensors are depicted. GNSS was used for satellite-based position determination, IMU for attitude determination, LiDAR and camera for generating a point cloud for path planning, and encoders of the drives for calculating odometry. The path planning is responsible for computing several tasks. Firstly, the optimal trajectories through the ridge of the compost windrows are calculated. Further, the computation of driving maneuvers necessary to approach the compost windrows and computation of turning maneuvers after the compost windrows have been turned, are performed.

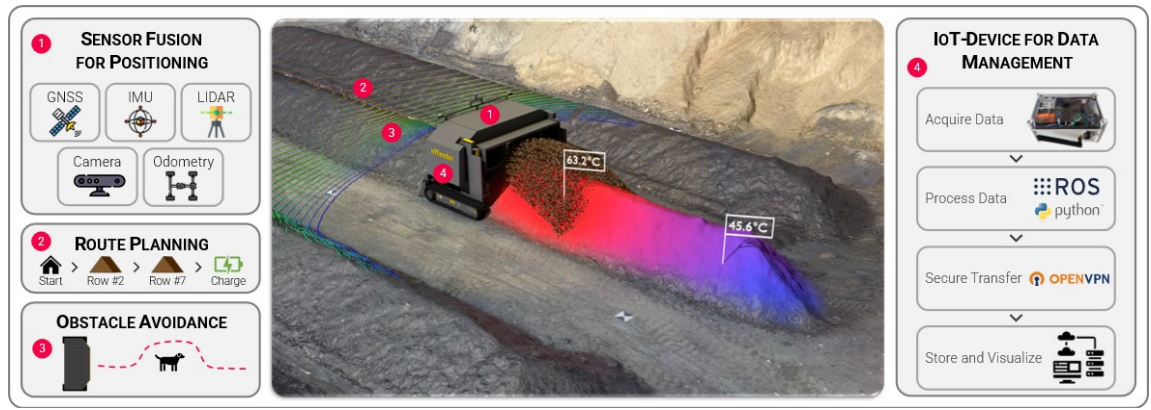

**Figure 22:** Overview of the process step and capabilities of the autonomous compost turner.

For the tasks of controlling the autonomous compost turner and the tasks of processing and web-based storage of necessary compost data, the IIoT module, whose methodological development was presented in [62], was employed in an industrial setting. Before the topic of systematic development with MBSE is taken up again, some aspects of the second research question should be discussed at this point.

# 5  Appendix

## 5.1  Author Contributions

Regarding the proportion of work per author, it should be noted that these aspects were specified according to CRediT standards [63] as follows:

Conceptualization, M.C.; methodology, M.C.; software, M.C.; validation, M.C.; investigation, M.C., E.R., F.T., C.S.; resources, M.C.; data curation, M.C.; writing - original draft preparation, M.C.; writing - review and editing, M.C.; visualization, M.C., F.T.; supervision, M.C.; project administration, M.C.; funding acquisition, M.C., E.R., F.T., C.S

## 5.2  Funding

Major parts of this research were funded by the TU Graz Open Access Publishing Fund and the Austrian Federal Ministry of Finance via the Austrian Research Promotion Agency in the research project ANTON (Grant No. 873670) [64], ANDREA (Grant No. 885368) [44] and CONCLUSION (Grant No. 900573) [65]. The projects used open-source software funded from the European Union's Horizon 2020 research and innovation programme under grant agreement No. 732287.